\documentclass[ijds,sglanonrev]{informs4}
\usepackage{enumitem}
\usepackage{eqndefns-left} 
\usepackage{booktabs,siunitx,threeparttable}
\usepackage{multirow}
\usepackage{threeparttable}
\usepackage{makecell}
\usepackage[font=small]{subcaption}
\usepackage{amsmath}
\usepackage{subcaption}
\RequirePackage{tgtermes}
\RequirePackage{newtxtext}
\RequirePackage{newtxmath}
\RequirePackage{bm}
\RequirePackage{endnotes}

\OneAndAHalfSpacedXII 

\usepackage{algorithm}
\usepackage{algpseudocode}
\usepackage{tikz}

\usepackage[colorlinks=true, citecolor=blue]{hyperref}

\usepackage{natbib}
 \bibpunct[, ]{(}{)}{,}{a}{}{,}%
 \def\bibfont{\small}%
\EquationsNumberedThrough    

\TheoremsNumberedThrough     
\ECRepeatTheorems  %

\MANUSCRIPTNO{IJDS-0001-2024.00}

\begin{document}


\RUNAUTHOR{Wen, Pinson, et al.}

\RUNTITLE{Probabilistic Wind Power Forecasting with Incomplete Data}

\TITLE{Marginalize, Rather than Impute: Probabilistic Wind Power Forecasting with Incomplete Data}

\ARTICLEAUTHORS{%
\AUTHOR{Honglin Wen}
\AFF{School of Electrical Engineering, Shanghai Jiao Tong University; Dyson School of Design Engineering, Imperial College London, \EMAIL{linlin00@sjtu.edu.cn}}

\AUTHOR{Pierre Pinson}
\AFF{Dyson School of Design Engineering, Imperial College London; Department of Technology, Management and Economics, Technical University of Denmark; Halfspace; Centre for Energy Research -- CoRE, Aarhus University, \EMAIL{p.pinson@imperial.ac.uk}}

\AUTHOR{Jie Gu, Zhijian Jin}
\AFF{School of Electrical Engineering, Shanghai Jiao Tong University, \EMAIL{gujie@sjtuu.edu.cn, zjjin@sjtu.edu.cn}}
} 

\ABSTRACT{%
Machine learning methods are widely and successfully used for probabilistic wind power forecasting, yet the pervasive issue of missing values (e.g., due to sensor faults or communication outages) has received limited attention. The prevailing practice is impute-then-predict, but conditioning on point imputations biases parameter estimates and fails to propagate uncertainty from missing features. Our approach treats missing features and forecast targets uniformly: we learn a joint generative model of features and targets from incomplete data and, at operational deployment, condition on the observed features and marginalize the unobserved ones to produce forecasts. This imputation-free procedure avoids error introduced by imputation and preserves uncertainty aroused from missing features. In experiments, it improves forecast quality in terms of continuous ranked probability score relative to impute-then-predict baselines while incurring substantially lower computational cost than common alternatives.
}%





\maketitle


\section{Introduction}\label{sec:Intro}

While renewable energy is widely recognized as a key driver of achieving carbon neutrality, its inherently limited predictability poses challenges for power system operations and electricity markets. Consequently, forecasting is considered an essential tool for system operators and has been evolving for decades, as highlighted in a recent review by \citet{hong2020energy}. Of particular interest is probabilistic forecasting \citep{pinson2007trading}, which leverages information such as numerical weather predictions and lagged observations up to the current time to communicate the probability of wind power generation at a future time in terms of densities, quantiles, and prediction intervals, etc.

With the rapid advancement of probabilistic forecasting, data availability, rather than just models, often limits forecast quality \citep{tawn2020missing}. In practice, the issues of missing values are prevalent in the energy sector. Missing values arise from sensor faults and communication interruptions \citep{liao2021data}, and are especially prevalent for assets exposed to harsh environments, such as offshore wind farms \citep{sun2021imputation}. Adversarial or accidental data losses can also occur through cyber incidents that suppress or remove measurements according to specific mechanisms \citep{pan2018cyber}. Beyond physical infrastructure, the lengthening data chain (spanning third-party providers, data sharing, and emerging markets \citep{pinson2022regression,falconer2025toward}) introduces latency and provision failures, increasing the incidence of incomplete data (i.e., missing values). 

Despite this reality, most existing studies implicitly assume complete data during both training and operational deployment, or address missingness with preprocessing methods, notably deletion and imputation. These remedies can degrade forecast quality by discarding information, biasing parameter estimates, and failing to propagate missing-feature uncertainty. For example, \citet{tawn2020missing} shows that in a dataset with 11.65\% missingness, list-wise deletion increases the normalized mean absolute error by 19\%, illustrating how naive preprocessing directly translates missingness into accuracy loss. Although list-wise deletion can bypass training issues from incomplete data, at operational deployment, missing features often render the forecasting model inapplicable, forcing operators to fall back on naive baselines such as persistence or climatology. Another intuitive idea is to sub-group the data and train a separate model for each missingness pattern \citep{biarosenbaum1984reducing}. However, this is computationally infeasible: with $d$-dimensional features, there are potentially up to $2^d$ distinct missingness patterns, and the data are fragmented across them, yielding severe sample inefficiency. In light of the above, it is prevailing to address incomplete data issues via imputation at both training and operational deployment, following an impute-then-predict pipeline reported in prior work \citep{liu2018wind, liu2020pv, liu2022missing}. \citet{srinivasan2025reduced} proposes a method that integrates sub-grouping with imputation. Specifically, the training data are grouped into overlapping sub-groups via clustering, after which missing values are imputed within each sub-group. However, as shown by \citet{josse2024consistency}, training forecasting models on imputed data—even when the imputations are statistically optimal—induces parameter-estimation bias, which in turn degrades forecast quality.

This gap motivates methods that natively learn from and predict with incomplete data, so a single model remains usable across arbitrary missingness patterns without manual reconfiguration or model switching \citep{bohlke2020resilient}. Work in this direction dates back at least to \citet{jones1980maximum}, \citet{harvey1984estimating}, and \citet{kohn1986estimation}, which recast ARMA/ARIMA models in state-space form and applied Kalman filtering with intermittent observations. That is, when an observation is missing, the measurement update is skipped, and only the state-update is performed. These approaches, however, are confined to linear models and primarily yield point forecasts. More recently, \citet{stratigakos2022towards} and \citet{xu2023availability} proposed robust-optimization and adversarial-training schemes that minimize a worst-case loss over prescribed missingness masks; their protection depends on the chosen uncertainty set/budget, and the estimators emphasize worst-case performance, so skillful predictive distributions are not obtained without extra modeling. In contrast, \citet{wen2024probabilistic} and \citet{stratigakos2025learning} make neural networks mask-aware by adding missingness indicators; however, the adaptation enters only through bias offsets rather than allowing the weights to adjust to masks. In addition, all of these methods are framed as regression models and, therefore, do not natively produce probabilistic scenarios or full predictive densities.

In this work, we aim to develop a single model that is estimated from incomplete datasets and, after operational deployment, issues probabilistic forecasts conditioned only on the features available at the time of issuance. Instead of using conditional distributional learning that embeds imputation, we design a joint modeling approach that learns the joint distribution of features and targets. Assuming the missing-data mechanism is missing at random (MAR), the missingness is ignorable (see Theorem \ref{theorem:inference}). Therefore, we can forgo explicit modeling of the missingness mechanism and instead maximize the observed-data likelihood to estimate the parameters of the data-distribution. During training, we fit the joint model by optimizing the observed-data likelihood; during operation, we produce forecasts by marginalizing over the unobserved inputs under the learned joint distribution.

Unlike \citet{wen2022wind} which employed a fully conditional specification by iteratively fitting one conditional model per feature, the present study learns the joint distribution of all features and the target within a single model. This shift eliminates the iterative per-feature fitting and yields substantially better computational efficiency during both training and operational deployment. Concretely, we use a variational autoencoder (VAE) \citep{kingma2013auto} with a neural-network based encoder and decoder. The decoder parameterizes a factorized Student’s-t likelihood over the concatenated feature–target vector (i.e., components are conditionally independent given the latent). For a partially observed sample, the likelihood is merely the product over the observed coordinates, which makes computation under missingness straightforward. Because incomplete data make the encoder’s posterior over latents more complex than in the fully observed case \citep{simkus2024improving}, we equip the encoder with a normalizing flow \citep{rezende2015variational} to obtain a more flexible and expressive approximation of the true posterior.

On an open dataset, the proposed method achieves consistent improvements in the continuous ranked probability score (CRPS) over the impute-then-predict baselines. It also maintains good calibration and reduces operational friction by forecasting directly from incomplete features.
In summary, our contributions are as follows:
\begin{enumerate}[label=(\arabic*)]
    \item Imputation-free joint modeling for probabilistic forecasting with incomplete data. We cast probabilistic forecasting with missing features as joint distribution learning and estimate parameters by maximizing the observed-data likelihood, yielding a single model that handles arbitrary missingness patterns at both training and test time without manual reconfiguration or model switching.
    \item Operational inference through marginalization. At operational deployment, the model conditions on observed features and marginalizes the unobserved ones to generate predictive scenarios, avoiding chained or iterative imputations and preserving uncertainty propagation.
    \item Flow-augmented VAE with a factorized decoder. We employ a VAE whose encoder uses normalizing flows for a flexible variational posterior under missingness, while the decoder specifies a conditionally independent Student’s-t likelihood over the feature–target vector, facilitating efficient evaluation of the observed-data likelihood.
\end{enumerate}

The structure of the paper is organized as follows. Section \ref{sec:Preliminary} introduces the foundational concepts of probabilistic forecasting, the mechanisms of missingness, and the impute-then-predict approach, while Section \ref{sec:Method} formulates the methodology and reveals the limitations of the impute-then-predict approach. Section \ref{sec:Model} details the design of the generative model, including model training and the forecasting algorithm. Next, Section \ref{sec:Case} outlines the experimental setups, benchmark models, and evaluation metrics. The results and discussion are found in Section \ref{sec:Results}. The paper concludes in Section \ref{sec:Conclusion}.

\section{Preliminaries}\label{sec:Preliminary}
Compared to point forecasting, which provides only the mean prediction of wind power generation, probabilistic forecasting communicates the full distributional information. This richer representation enables more informed decision-making under uncertainty \citep{morales2013integrating}. Accordingly, this work primarily focuses on probabilistic forecasting.
This section covers the fundamentals of probabilistic wind power forecasting, outlines the definitions of missingness mechanisms, and describes the impute-then-predict approach.

\subsection{Probabilistic Wind Power Forecasting}

For simplicity, let us consider univariate wind power forecasting, with a special focus on very short-term cases,\endnote{We note that the method designed in this work is also applicable to general cases, such as multivariate forecasting and short-term forecasting.} where missingness often occurs due to sensor failures and communication outages. Let $Y_t$ denote the random variable for the wind power generation value at time $t$, and $y_t$ its realization. Probabilistic wind power forecasting aims to communicate the probability of wind power generation at time $t+k$ (also referred to as targets) with the information up to time $t$, denoted as $\Omega_t$ (also referred to as features), based on model $\mathcal{M}_{\boldsymbol{\psi}}$ with parameters $\boldsymbol{\psi}$.
Probabilistic forecasting is described as
\begin{equation}
\label{eq:pwpf}
    \hat{p}_{Y_{t+k}}(y_{t+k})= p(y_{t+k}\mid\Omega_t;\boldsymbol{\psi}).
\end{equation}
The model $\mathcal{M}_{\boldsymbol{\psi}}$ could be either parametric or non-parametric (with respect to distributional assumptions). The parametric method depends on assumed distributions such as Gaussian, Beta, and Logit-normal, with the parameters of these shapes being determined using statistical and machine learning techniques \citep{pinson2012very, wen2022sparse}. Non-parametric methods are distribution-free and can be developed using quantile regression (QR) \citep{wan2016direct}, normalizing flow \citep{wen2022continuous}, diffusion models \citep{liao2021windgmmn}, etc.
In very short-term probabilistic forecasting, the information $\Omega_t$ is often composed of previous values of length $h$, i.e., $y_{t-h+1},y_{t-h+2},\cdots,y_{t-1},y_t$. For convenience, we write them as $ \mathbf{y}_t $ and the corresponding random variable as $\mathbf{Y}_t$.
In particular, we consider the underlying process to be stationary. As a result, the parameters of the model can be estimated using the historical dataset.

\subsection{Missingness Mechanism}

Let the random variable $M_t$ represent whether $y_t$ is missing, with $m_t\in \{0,1 \}$ being its realization. $m_t=1$ means $y_t$ is missing, whereas $m_t=0$ means $y_t$ is observed. Accordingly, the corresponding mask variable for the variable $\mathbf{Y}_t$ is written  as $\mathbf{M}_t$, whose realization is $\mathbf{m}_t\in \{0,1\}^h$. We specifically refer to $\mathbf{m}_t$ as the mask corresponding to $\mathbf{y}_t$. The data sample can be split into an observed part $\mathbf{y}_t^o$ and a missing part $\mathbf{y}_t^m$, i.e., $\mathbf{y}_t = (\mathbf{y}_t^o,\mathbf{y}_t^m)$. In many cases, the mechanism of missing data is characterized by the joint distribution of both the data and its mask. Employing the selection model \citep{little2019statistical}, this joint distribution can be expressed as the factorization:
\begin{equation}
    p(\mathbf{y}_t,\mathbf{m}_t;\boldsymbol{\theta},\boldsymbol{\gamma})=p(\mathbf{y}_t;\boldsymbol{\theta})p(\mathbf{m}_t\mid\mathbf{y}_t;\boldsymbol{\gamma}),
\end{equation}
with $\boldsymbol{\theta}$ and $\boldsymbol{\gamma}$ denoting the parameters for the data distribution and the mask distribution, respectively.

\begin{figure}[ht]
    \centering
    \begin{subfigure}[b]{0.45\textwidth}
        \centering
        \includegraphics[width=0.95\textwidth]{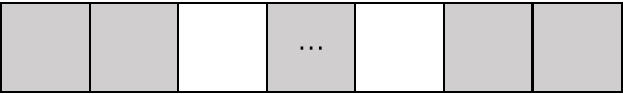}
        \caption{Data are missing sporadically}
        \label{fig:sub1}
    \end{subfigure}
    \hfill 
    \begin{subfigure}[b]{0.45\textwidth}
        \centering
        \includegraphics[width=0.95\textwidth]{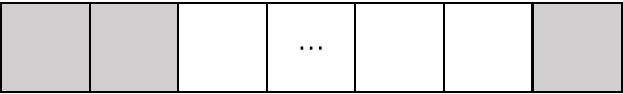}
        \caption{Data are missing in blocks}
        \label{fig:sub2}
    \end{subfigure}
    
    \caption{Illustrative samples with missingness}
    \label{fig:missingness}
\end{figure}

In modern statistical theory \citep{little2019statistical}, missingness mechanisms can be classified into three categories: missing completely at random (MCAR), missing at random (MAR), and missing not at random (MNAR). We introduce the definitions in what follows.
\begin{definition}[MCAR]
The data are missing completely at random if 
\[ p(\mathbf{m}_t\mid\mathbf{y}_t;\boldsymbol{\gamma})=p(\mathbf{m}_t;\boldsymbol{\gamma}).
\]
\end{definition}
\noindent The MCAR mechanism entails that the missingness is independent of the data.
\begin{definition}[MAR]
The data are missing at random if 
\[ p(\mathbf{m}_t\mid\mathbf{y}_t;\boldsymbol{\gamma})=p(\mathbf{m}_t\mid\mathbf{y}_t^o;\boldsymbol{\gamma}).
\]
\end{definition}
\noindent The MAR mechanism entails that missingness may depend on observed values but not on the missing ones, indicating that the likelihood of missingness remains unchanged regardless of the values of the missing data. Particularly, MCAR is a special case of MAR.
\begin{definition}[MNAR]
The data are missing not at random if 
\[p(\mathbf{m}_t\mid\mathbf{y}_t;\boldsymbol{\gamma})=p(\mathbf{m}_t\mid\mathbf{y}_t^o,\mathbf{y}_t^m;\boldsymbol{\gamma}).
\]
\end{definition}
\noindent The MNAR mechanism entails that missingness depends on the unobserved values themselves. Here, we assume that wind power data are MAR (including MCAR), as the missingness is often caused by sensor failures and communication errors, which are irrelevant to the sample values. Although the mechanism may be MNAR due to cyber attacks, as noted by \citet{xu2023availability}, we do not address this issue in the current work and leave it as a direction for future research.
\begin{remark}
It is worth noting that the missingness mechanism is irrelevant to how missing values are distributed within the dataset. For example, missing values occurring in blocks or sporadically, as shown in Figure~\ref{fig:missingness} (where gray boxes indicate observed values and white boxes indicate unobserved values) can both be attributed to the MAR mechanism, provided that the missingness is unrelated to the values. Conversely, both scenarios can be attributed to the MNAR mechanism if the missingness is dependent on the values (for instance, data are missing when their values exceed a threshold).    
\end{remark}

\subsection{Impute-then-predict Approach}
In an autoregressive setting, both the features $\mathbf{y}_{t}$ and the targets $y_{t+k}$ can be missing. This complicates the training of model \eqref{eq:pwpf}: missing values block the forward pass and, hence, the gradient computation. From the historical sample $\{(\mathbf{y}_t,y_{t+k})\mid t=1,2,\cdots,T \}$, we therefore restrict estimation to timestamps with observed targets $y_{t+k}$. We denote the indices of these instances as $\mathcal{T}$; thus, our dataset becomes $\{(\mathbf{y}_t,y_{t+k})\mid t\in \mathcal{T} \}$. With $\mathbf{y}_t = (\mathbf{y}_t^o,\mathbf{y}_t^m)$, a common workaround is to impute the missing features prior to estimation and forecasting using an imputer $f_\text{IM}$. For point imputation, the completed feature vector is
\begin{equation}
    \hat{\mathbf{y}}_t=(\mathbf{1}-\mathbf{m}_t)\odot \mathbf{y}_t+\mathbf{m}_t\odot f_\text{IM}(\mathbf{y}_t^o,\mathbf{m}_t),
\end{equation}
where $\odot$ is the elementwise multiplication operator.
Estimation under the impute-then-predict pipeline proceeds by maximizing (or minimizing a loss based on) the conditional likelihood given the imputed features,
\begin{equation}
\max_{\boldsymbol{\psi}} \sum_{t\in \mathcal{T}}\log p(y_{t+k}\mid\hat{\mathbf{y}_t};\boldsymbol{\psi}).
\end{equation}
Recently, many works have adopted on-the-fly imputation inside the impute-then-forecast pipeline \citep{salinas2020deepar}. Some go further by jointly learning the imputer and the forecaster during training \citep{liu2020pv,liu2022missing,meng2025probabilistic}.
Nevertheless, as \citet{josse2024consistency} demonstrates, training on point imputations—even optimal ones—induces estimation bias. We later contrast this with observed-data likelihood maximization, which integrates over $\mathbf{y}_t^m$ instead of conditioning on point imputations.

\section{Methodology}\label{sec:Method}

\begin{figure*}[!ht]
\centering
\includegraphics[width=0.75\textwidth]{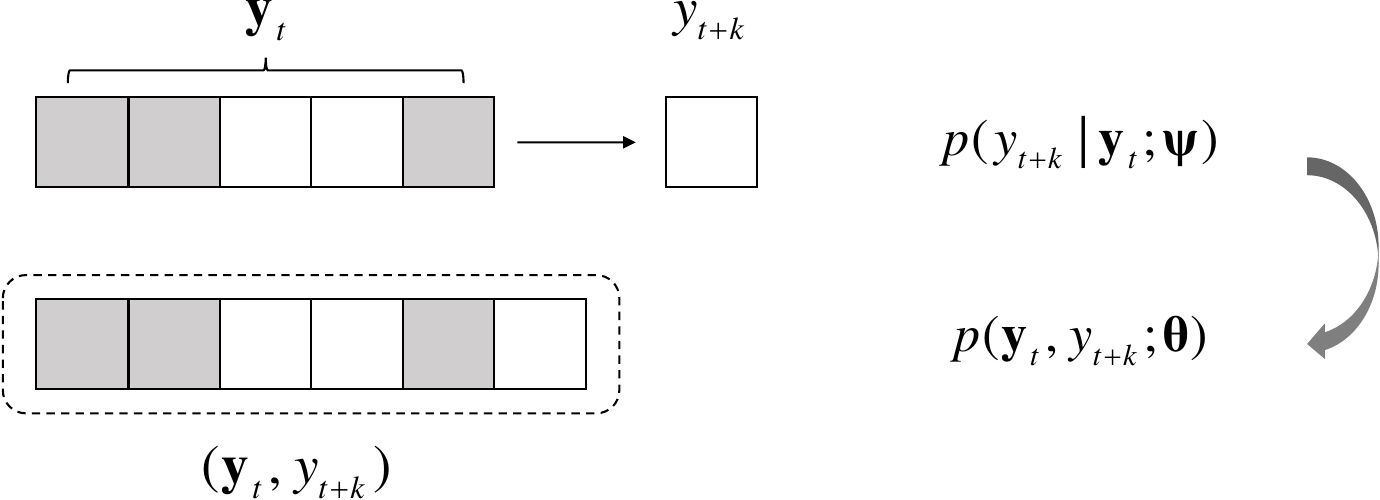}
\caption{Transition from conditional distribution modeling to joint distribution modeling, where gray blocks indicate observed values and white blocks indicate unobserved values.}
\label{fig:framework}
\end{figure*}

For each lead time $k$, if the joint distribution over current features and future targets is available, $p(\mathbf{y}_t,y_{t+k};\boldsymbol{\theta})$ with $\mathbf{y}_t=(\mathbf{y}_t^o,\mathbf{y}_t^m)$, then the predictive conditional, given the observed features, is obtained by marginalization:

\begin{equation}
\begin{aligned}
p(y_{t+k}\mid \mathbf{y}_t^{o};\boldsymbol{\theta})
&= \int_{\mathbf{y}_t^{m}}
   p(y_{t+k}\mid \mathbf{y}_t^{o},\mathbf{y}_t^{m};\boldsymbol{\theta})\\[-2pt]
&\qquad \times
   p(\mathbf{y}_t^{m}\mid \mathbf{y}_t^{o};\boldsymbol{\theta})\,
   d\mathbf{y}_t^{m}\\[2pt]
&=
   \frac{p(\mathbf{y}_t^{o},y_{t+k};\boldsymbol{\theta})}
        {p(\mathbf{y}_t^{o};\boldsymbol{\theta})}.
\end{aligned}
\end{equation}
where
\begin{equation*}
\begin{aligned}
    &p(\mathbf{y}_t^o,y_{t+k};\boldsymbol{\theta})=\int_{\mathbf{y}_t^m} p(\mathbf{y}_t^o,\mathbf{y}_t^m,y_{t+k})d\mathbf{y}_t^m,\\
    &p(\mathbf{y}_t^o;\boldsymbol{\theta})=\int_{\mathbf{y}_t^m} \int_{y_{t+k}} p(\mathbf{y}_t^o,\mathbf{y}_t^m,y_{t+k})d\mathbf{y}_t^m dy_{t+k}.
\end{aligned}
\end{equation*}
Thus, by modeling the joint $p(\mathbf{y}_t,y_{t+k};\boldsymbol{\theta})$ with a parametric model $\mathcal{M}_{\boldsymbol{\theta}}$, we can derive the needed conditional $p(y_{t+k}\mid\mathbf{y}_t^o;\boldsymbol{\theta})$ via marginalization over the missing features $\mathbf{y}_t^m$. In this view, forecasting reduces to learning a joint distribution from incomplete data and, at operation time, evaluating the conditional by integrating out the unobserved features, e.g., by sampling depending on $\mathcal{M}_{\boldsymbol{\theta}}$. We illustrate the model framework in Figure~\ref{fig:framework}.
\begin{remark}
Under the above framework, the limitation of the impute-then-predict method is immediate. The approach replaces the unknown $\mathbf{y}^m_t$ with a single imputed value $\hat{\mathbf{y}}_t^m$, and then evaluates $p(y_{t+k}\mid\mathbf{y}_t^o,\hat{\mathbf{y}}_t^m)$. This treats the imputed values as error-free, so uncertainty about the missing features does not propagate into the forecasts. For rigorous proof, readers can refer to \citet{josse2024consistency}.
\end{remark}

For notational convenience, define the concatenated random vector
\begin{equation*}
\mathbf{Z}_t = \begin{bmatrix}\mathbf{Y}_t \\[2pt] Y_{t+k}\end{bmatrix}
\in \mathbb{R}^{h+1}, \qquad
\mathbf{z}_t = \begin{bmatrix}\mathbf{y}_t \\[2pt] y_{t+k}\end{bmatrix}
\in \mathbb{R}^{h+1},
\end{equation*}
where $h\!=\!\dim(\mathbf{Y}_t)$.  With slight abuse of notation, we still denote the mask variable for the variable $\mathbf{Z}_t$ as $\mathbf{M}_t$, whose realization is $\mathbf{m}_t\in \{0,1\}^{h+1}$. The data sample can be split into an observed part $\mathbf{z}_t^o$ and a missing part $\mathbf{z}_t^m$, i.e., $\mathbf{z}_t = (\mathbf{z}_t^o,\mathbf{z}_t^m)$.
We now recall the classical identifiability result for estimation with incomplete data established by \citet{rubin1976inference}, and include a self-contained proof for completeness.
\begin{theorem}[Ignorability under MAR \citep{rubin1976inference}]
\label{theorem:inference}
    Let $\boldsymbol{\gamma}$ such that for all $t$, $p(\mathbf{m}_t\mid\mathbf{z}_t;\boldsymbol{\gamma})>0$. Assuming data are MAR, $p(\mathbf{z}_t^o,\mathbf{m}_t;\boldsymbol{\theta},\boldsymbol{\gamma})$ is proportional to $p(\mathbf{z}_t^o,\mathbf{m}_t;\boldsymbol{\theta})$  w.r.t. $\boldsymbol{\theta}$, so that inference for $\boldsymbol{\theta}$ can be obtained by maximizing the likelihood $p(\mathbf{z}_t^o,\mathbf{m}_t;\boldsymbol{\theta})$ while ignoring the missingness mechanism.
\end{theorem}
\begin{proof}{Proof}
The likelihood of observation can be derived by integrating over missing variables, i.e.,
\begin{equation*}
\begin{aligned}
p(\mathbf{z}_t^o,\mathbf{m}_t;\boldsymbol{\theta},\boldsymbol{\gamma})
&= \int_{\mathbf{z}_t^m}
   p(\mathbf{z}_t^o,\mathbf{z}_t^m;\boldsymbol{\theta}) \\
&\quad\times
   p(\mathbf{m}_t\mid\mathbf{z}_t^o,\mathbf{z}_t^m;\boldsymbol{\gamma})
   \, d\mathbf{z}_t^m .
\end{aligned}
\end{equation*}
In cases where data are MAR, the expression $p(\mathbf{m}_t\mid\mathbf{z}_t^o,\mathbf{z}_t^m;\boldsymbol{\gamma})$ simplifies to $p(\mathbf{m}_t\mid\mathbf{z}_t^o;\boldsymbol{\gamma})$. This allows it to be extracted from the integral, i.e.,
\begin{equation*}
\begin{aligned}
p(\mathbf z_t^o,\mathbf m_t;\boldsymbol\theta,\boldsymbol\gamma)
&= \int_{\mathbf z_t^m}
   p(\mathbf z_t^o,\mathbf z_t^m;\boldsymbol\theta)\,
   d\mathbf z_t^m \\
&\quad\times
   p(\mathbf m_t \mid \mathbf z_t^o;\boldsymbol\gamma).
\end{aligned}
\end{equation*}
As we are interested in $\boldsymbol{\theta}$, we have
\begin{equation*}
p(\mathbf{z}_t^o,\mathbf{m}_t;\boldsymbol{\theta},\boldsymbol{\gamma})\propto \int_{\mathbf{z}_t^m} p(\mathbf{z}_t^o,\mathbf{z}_t^m;\boldsymbol{\theta})d \mathbf{z}_t^m,
\end{equation*}
which means the observed-data likelihood is independent of the probability of missingness. \Halmos
\end{proof}
Consequently, MAR (including MCAR) is termed ignorable because the missingness mechanism need not be modeled for inference. In particular, Theorem~\ref{theorem:inference} implies that, under MAR settings, $\boldsymbol{\theta}$ can be estimated by maximizing the observed-data likelihood of $p(\mathbf{z}_t;\boldsymbol{\theta})$, even when some components of $\mathbf{z}_t$ are missing. 
Given $T$ samples, $\{\mathbf{z}_1, \mathbf{z}_2,\cdots, \mathbf{z}_T \}$, the log-likelihood can be derived as
\begin{equation*}
\mathcal{L}(\boldsymbol{\theta};\mathbf{z}_t^o)=\frac{1}{T}\sum_{t=1}^T\log\int_{\mathbf{z}_t^m} p(\mathbf{z}_t^o,\mathbf{z}_t^m;\boldsymbol{\theta})d \mathbf{z}_t^m.
\end{equation*}
The parameters $\boldsymbol{\theta}$ are estimated via machine learning using (variational) maximum likelihood, and the estimates are denoted by $\hat{\boldsymbol{\theta}}$.

While at the forecasting stage, we derive the conditional distribution of $\mathbf{z}_t^m$ by marginalizing the learned joint distribution $p(\mathbf{z}_t;\hat{\boldsymbol{\theta}})$, i.e.,
\begin{equation*}
    p(\mathbf{z}_t^m\mid\mathbf{z}_t^o;\hat{\boldsymbol{\theta}})=\frac{p(\mathbf{z}_t^m,\mathbf{z}_t^o;\hat{\boldsymbol{\theta}})}{\int_{\mathbf{z}_t^m}p(\mathbf{z}_t^m,\mathbf{z}_t^o;\hat{\boldsymbol{\theta}}) d\mathbf{z}_t^m}.
\end{equation*}
As $\mathbf{z}_t^m$ is composed of $\mathbf{y}_t^m$ and $y_{t+k}$, i.e., $\mathbf{z}_t^m=[{\mathbf{y}_t^m}^\top, y_{t+k}]^\top$, the probabilistic forecast for $y_{t+k}$ can be derived via marginalization over $\mathbf{y}_t^m$:
\begin{equation}
    p(y_{t+k}\mid\mathbf{z}_t^o;\hat{\boldsymbol{\theta}})=\int_{\mathbf{y}_t^m} p(\mathbf{z}_t^m\mid\mathbf{z}_t^o;\hat{\boldsymbol{\theta}}) d\mathbf{y}_t^m.
\end{equation}

In summary, modeling the joint distribution of both features and targets enables us to obtain probabilistic predictions from observations, specifically $p(y_{t+k}\mid\mathbf{y}_t^o)$ as $\mathbf{y}_t^o=\mathbf{z}_t^o$. This method circumvents the possible bias associated with imputation and necessitates the creation of just a single model, a process that is computationally feasible.
In the joint modeling framework, during model estimation, a key challenge involves learning the distribution $p(\mathbf{z}_t;\hat{\boldsymbol{\theta}})$ using incomplete data. Meanwhile, at the forecasting stage, efficiently computing marginalization presents its own difficulties. For that, we design a generative model for joint distribution modeling, which is detailed in the next section.

\section{Proposed Model}\label{sec:Model}
Our goal is to learn a generative model whose samples are consistent with incomplete observations. During training, we handle incomplete data by maximizing an importance-weighted evidence lower bound (ELBO) (the IWAE objective of \citet{mattei2019miwae}) to estimate the model parameters. Because the decoder assumes conditionally independent outputs, the observed-data likelihood is easy to evaluate given the latent variables (we simply mask out the missing coordinates).
At operational deployment, the incomplete input is passed through the encoder to draw latent samples; each latent is then ancestrally decoded to produce a proposal scenario. Repeating this yields a bank of proposals, which we reweight by their observed-data likelihood (importance sampling) and resample to obtain calibrated predictive scenarios. The probabilistic forecast of the target is thus obtained by marginalizing over the missing features using these resampled trajectories.
Below, we first describe the model architecture, then the estimation procedure, and finally how the fitted model is used for operational forecasting.

\subsection{Model Architecture}

In particular, our model is constructed using the variational auto-encoder (VAE) framework \citep{kingma2013auto}, which posits that data are derived from a latent variable $\mathbf{u}_t\in \mathbb{R}^r$ following the distribution $ p(\mathbf{u}_t)$, where $r$ represents the dimensionality. Specifically, observations are generated through a \textit{decoder} model $p(\mathbf{z}_t\mid\mathbf{u}_t;\boldsymbol{\theta})$. The data generation process is characterized as
\begin{equation}
    \mathbf{u}_t\sim p(\mathbf{u}_t), \ \mathbf{z}_t \sim p(\mathbf{z}_t\mid\mathbf{u}_t;\boldsymbol{\theta}).
\end{equation}
Then, the log-likelihood of the observations can be specified as
\begin{equation}
\label{eq:vae}    \log p(\mathbf{z}_t;\boldsymbol{\theta})=\log\int_{\mathbf{u}_t}p(\mathbf{z}_t\mid\mathbf{u}_t;\boldsymbol{\theta})p(\mathbf{u}_t)d\mathbf{u}_t.
\end{equation}
Typically, the prior distribution for the latent variable is assumed to be a normal distribution, expressed as
\begin{equation*}
    p(\mathbf{u}_t) = \mathcal{N}(\mathbf{u}_t \mid \mathbf{0}, \mathbf{I}),
\end{equation*}
where $\mathbf{0} \in \mathbb{R}^r$ denotes a zero vector, and $\mathbf{I} \in \mathbb{R}^{r \times r}$ is a diagonal identity matrix.

\begin{figure}[!ht]
\centering
\includegraphics[width=0.4\textwidth]{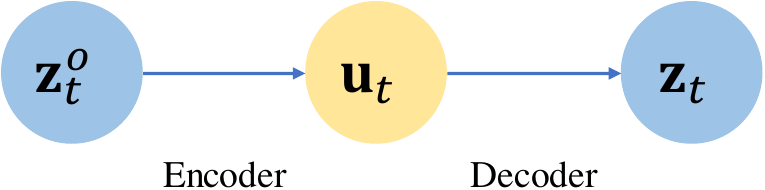}
\caption{Illustration of how complete samples are predicted by using the encoder and decoder models.}
\label{fig:workflow}
\end{figure}

Our inference task is to approximate the latent posterior from the \emph{observed} components only, $q(\mathbf{u}_t\mid\mathbf{z}_t^o;\boldsymbol{\phi})$, and then predict the full sample via the decoder $p(\mathbf{z}_t\mid\mathbf{u}_t;\boldsymbol{\theta})$; see Figure~\ref{fig:workflow} for the graphical model. Given \(q(\mathbf{u}_t \mid \mathbf{z}_t^{o};\boldsymbol{\phi})\) and \(p(\mathbf{z}_t \mid \mathbf{u}_t; \boldsymbol{\theta})\), the conditional for the missing part is obtained through marginalization,
\begin{equation}
\begin{aligned}
p(\mathbf{z}_t^{m}\mid \mathbf{z}_t^{o};\boldsymbol{\theta},\boldsymbol{\phi})
&= \int_{\mathbf{u}_t}
   p(\mathbf{z}_t^{m}\mid \mathbf{z}_t^{o},\mathbf{u}_t;\boldsymbol{\theta}) \\
&\quad\times
   q(\mathbf{u}_t \mid \mathbf{z}_t^{o};\boldsymbol{\phi})\, d\mathbf{u}_t .
\end{aligned}
\end{equation}
which we evaluate by sampling from \(q(\mathbf{u}_t \mid \mathbf{z}_t^{o};\boldsymbol{\phi})\) during operation.

For computation, we feed the encoder with a representation of the partially observed sample. Particularly, we employ a function $g$ to map $\mathbf{z}_t$ to a value within the Real number domain $\mathbb{R}^{h+1}$ and parameterize the encoder as \(q(\mathbf{u}_t \mid g(\mathbf{z}_t);\boldsymbol{\phi})\).\endnote{In implementation, the encoder takes the masked embedding $g(\mathbf{z}_t)$ as input; for clarity in the model description that follows, we write this as $\mathbf{z}_t^o$.} A simple and effective choice is to zero-fill the missing entries
\[
g(\mathbf{z}_t) =  (\mathbf{1}-\mathbf{m}_t)\odot \mathbf{z}_t  \in \mathbb{R}^{h+1}.
\]
The subsequent sections detail the design of the encoder and decoder utilized in this study.

\subsubsection{Decoder}

We model the decoder with a factorized Student’s-$t$ likelihood:
\begin{equation}
\begin{aligned}
p(\mathbf{z}_t \mid \mathbf{u}_t;\boldsymbol{\theta})
&= \prod_{j=1}^{h+1}
   \mathrm{St}\!\big(
      z_{t,j}\,\big|\,
      \mu_j(\mathbf{u}_t;\boldsymbol{\theta}), \\
&\qquad\qquad\quad
      \sigma_j(\mathbf{u}_t;\boldsymbol{\theta}),
      \nu_j(\mathbf{u}_t;\boldsymbol{\theta})
   \big).
\end{aligned}
\end{equation}
where each component is a univariate Student’s-$t$ with location $\mu_j$, scale $\sigma_j>0$, and degrees of freedom $\nu_j>0$ produced by neural networks. 
With the mask $\mathbf{m}_t\in\{0,1\}^{h+1}$, we define the observed index set
$\mathcal{O}_t=\{j: m_{t,j}=0\}$. Under the factorization above, the observed–data likelihood is
\begin{equation*}
\begin{aligned}
p(\mathbf{z}_t^{o}\mid \mathbf{u}_t;\boldsymbol{\theta})
&= \prod_{j\in\mathcal{O}_t}
   \mathrm{St}\!\big(
      z_{t,j}\,\big|\,
      \mu_j(\mathbf{u}_t;\boldsymbol{\theta}),\\
&\qquad\qquad
      \sigma_j(\mathbf{u}_t;\boldsymbol{\theta}),
      \nu_j(\mathbf{u}_t;\boldsymbol{\theta})
   \big).
\end{aligned}
\end{equation*}
and the corresponding log–likelihood is the masked sum
\begin{equation*}
\begin{aligned}
\log p(\mathbf{z}_t^{o}\mid \mathbf{u}_t;\boldsymbol{\theta})
&= \sum_{j\in\mathcal{O}_t}
   \log \mathrm{St}\!\big(
      z_{t,j}\,\big|\,
      \mu_j(\mathbf{u}_t;\boldsymbol{\theta}), \\
&\qquad\qquad\quad
      \sigma_j(\mathbf{u}_t;\boldsymbol{\theta}),
      \nu_j(\mathbf{u}_t;\boldsymbol{\theta})
   \big).
\end{aligned}
\end{equation*}
The factorized form corresponds to a diagonal covariance given the latent $\mathbf{u}_t$; thus, the likelihood of a partially observed sample is simply the product over the observed coordinates, enabling straightforward computation under missingness.

\subsubsection{Encoder}
Because missing data make the encoder’s posterior \(q(\mathbf{u}_t\mid \mathbf{z}_t^{o})\) complex and non-Gaussian \citep{simkus2024improving}, we parameterize it as a normalizing flow that transforms a simple base distribution. 
Let the base distribution be Gaussian and denote its law by:
\[
\mathbf{u}_t^{(0)} \sim \mathcal{N}\!\big(\boldsymbol{\mu}_0(\mathbf{z}_t^{o};\boldsymbol{\phi}),\,\boldsymbol{\Sigma}_0(\mathbf{z}_t^{o};\boldsymbol{\phi})\big),
\]
where \((\boldsymbol{\mu}_0,\boldsymbol{\Sigma}_0)\) is produced by the encoder network from the observed inputs. We then apply a sequence of invertible, differentiable transforms \(\{f_n\}_{n=1}^N\). For clarity, we reserve the superscript \((0)\) for the base variable \(\mathbf{u}_t^{(0)}\) and write \(\mathbf{u}_t=\mathbf{u}_t^{(N)}\) for the final latent. The transforms operate as:
\begin{equation}
\begin{aligned}
\mathbf{u}_t^{(n)} &= f_n\!\big(\mathbf{u}_t^{(n-1)},\mathbf{z}_t^{o};\,\boldsymbol{\phi}\big),
\qquad n=1,\dots,N,\\
\mathbf{u}_t      &= \mathbf{u}_t^{(N)}.
\end{aligned}
\end{equation}
The resulting posterior likelihood follows from the change of variables:
\begin{equation}
\begin{aligned}
\log q(\mathbf{u}_t \mid \mathbf{z}_t^{o};\boldsymbol{\phi})
&= \log \mathcal{N}\!\big(\mathbf{u}_t^{(0)};\boldsymbol{\mu}_0,\boldsymbol{\Sigma}_0\big) \\
&\quad - \sum_{n=1}^{N} 
   \log \left|
        \det \frac{\partial f_n}{\partial \mathbf{u}_t^{(n-1)}}
   \right|.
\end{aligned}
\end{equation}
Following \citet{wen2022continuous}, we employ autoregressive normalizing flows with tractable Jacobians, enabling stable and efficient training. Further details are provided in the survey by \citet{papamakarios2021normalizing}.

\subsection{Model Training}

As discussed in Section~\ref{sec:Method}, under MAR missingness, the likelihood involves only the observed components. 
We therefore estimate $\boldsymbol{\theta}$ by maximizing the \emph{observed-data} log-likelihood:
\begin{equation*}
    \hat{\boldsymbol{\theta}} = \arg \max \limits_{\boldsymbol{\theta}} \sum_{t=1}^T  \log p(\mathbf{z}_t^o;\boldsymbol{\theta}).
\end{equation*}
For the used generative model, we derive the log-likelihood using the chain rule:
\begin{equation}
    \log p(\mathbf{z}_t^o;\boldsymbol{\theta}) = \ \sum_{t=1}^T  \log \int_{\mathbf{u}_t}  p(\mathbf{z}_t^o\mid\mathbf{u}_t;\boldsymbol{\theta})p(\mathbf{u}_t)d\mathbf{u}_t.
\end{equation}
Similar to the complete data cases, we rewrite the log-likelihood with the help of the variational distribution as
\begin{equation}\label{eq:likelihood}
\begin{aligned} 
    &\sum_{t=1}^T  \log \int_{\mathbf{u}_t}  \frac{p(\mathbf{z}_t^o\mid\mathbf{u}_t;\boldsymbol{\theta})p(\mathbf{u}_t)}{q(\mathbf{u}_t\mid \mathbf{z}_t^o;\boldsymbol{\phi})}q(\mathbf{u}_t\mid \mathbf{z}_t^o;\boldsymbol{\phi}) d\mathbf{u}_t \\
    =&\sum_{t=1}^T  \log \mathbb{E}_{q(\mathbf{u}_t\mid \mathbf{z}_t^o);\boldsymbol{\phi})}\left[\frac{p(\mathbf{z}_t^o\mid\mathbf{u}_t;\boldsymbol{\theta})p(\mathbf{u}_t)}{q(\mathbf{u}_t\mid \mathbf{z}_t^o);\boldsymbol{\phi})}\right].
\end{aligned}
\end{equation}

Since \eqref{eq:likelihood} is generally intractable, we apply Jensen’s inequality to obtain the
\emph{evidence lower bound} (ELBO):
\begin{equation}
\begin{aligned}
\sum_{t=1}^T  & \log p(\mathbf{z}_t^o;\boldsymbol{\theta}) 
\ge \sum_{t=1}^T \mathbb{E}_{q} \Bigg[
    \log \frac{
        p(\mathbf{z}_t^o\mid\mathbf{u}_t;\boldsymbol{\theta})\,
        p(\mathbf{u}_t)}
        {q(\mathbf{u}_t\mid \mathbf{z}_t^o;\boldsymbol{\phi})}
\Bigg] \\
&= \sum_{t=1}^T \mathbb{E}_{q} \Big[
    \log p(\mathbf{z}_t^o\mid\mathbf{u}_t;\boldsymbol{\theta})
    + \log p(\mathbf{u}_t)\\
&\qquad\qquad
    - \log q(\mathbf{u}_t\mid \mathbf{z}_t^o;\boldsymbol{\phi})
\Big] \\
&=: \mathcal{L}_{\rm ELBO}(\boldsymbol{\theta},\boldsymbol{\phi}).
\end{aligned}
\end{equation}
To tighten the bound, we use the importance-weighted auto-encoder (IWAE) objective, drawing $S$ samples $\{\mathbf{u}_{t,i} \}_{i=1}^S$ from the variational posterior $q(\mathbf{u}_t\mid \mathbf{z}_t^o;\boldsymbol{\phi})$, i.e.,
\begin{equation}
\label{eq:elbo}
\begin{aligned}
\mathcal{L}_{\text{IWAE}}(\boldsymbol{\theta},\boldsymbol{\phi})
&= \sum_{t=1}^T  
   \log \left[ \frac{1}{S}\sum_{i=1}^S l_{t,i} \right], \\[2pt]
l_{t,i}
&:= \frac{
      p(\mathbf{z}_t^o\mid\mathbf{u}_{t,i};\boldsymbol{\theta})\,
      p(\mathbf{u}_{t,i})
    }{
      q(\mathbf{u}_{t,i}\mid \mathbf{z}_t^o;\boldsymbol{\phi})
    }.
\end{aligned}
\end{equation}
Consequently, we estimate the parameters by maximizing $\mathcal{L}_{\text{IWAE}}(\boldsymbol{\theta},\boldsymbol{\phi})$, and denote the final estimate as $\hat{\boldsymbol{\theta}}$, $\hat{\boldsymbol{\phi}}$.

\subsection{Forecasting Algorithm}

At the forecasting stage, $y_{t+k}$ is inherently missing; that is, $\mathbf{z}_t^m=\big[\,{\mathbf{y}_t^{m}}^\top,\; y_{t+k}\,\big]^\top$. Following \citet{rezende2014stochastic}, the underlying conditional distribution of the missing variables, given the observations, is
\begin{equation*}
\begin{aligned}
p(\mathbf{z}_t^m\mid\mathbf{z}_t^o;\boldsymbol{\theta})
&= \int_{\mathbf{u}_t}
   p(\mathbf{z}_t^m\mid\mathbf{z}_t^o,\mathbf{u}_t;\boldsymbol{\theta}) \\
&\quad\times
   p(\mathbf{u}_t\mid\mathbf{z}_t^o;\boldsymbol{\theta})\, d\mathbf{u}_t,
\end{aligned}
\end{equation*}
where we approximate by replacing $p(\mathbf{u}_t\mid\mathbf{z}_t^o;\boldsymbol{\theta})$ with the variational posterior $q(\mathbf{u}_t\mid\mathbf{z}_t^o;\boldsymbol{\phi})$.
Because the variational posterior $q(\mathbf{u}_t\mid \mathbf{z}_t^o;\hat{\boldsymbol{\phi}}))$ is only an approximation of the true posterior $p(\mathbf{u}_t\mid\mathbf{z}_t^o;\boldsymbol{\theta})$, direct ancestral sampling can misrepresent $p(\mathbf{z}_t^m\mid\mathbf{z}_t^o;\boldsymbol{\theta})$. Therefore, we use importance resampling \citep{chopin2020introduction}. Treat $q(\mathbf{u}_t\mid \mathbf{z}_t^o;\hat{\boldsymbol{\phi}})$ as a proposal distribution, draw the sample $\{\mathbf{u}_{t,i}\}_{i=1}^S$, and for each $\mathbf{u}_{t,i}$ draw $\mathbf{z}_{t,i}\sim p(\mathbf{z}_t \mid \mathbf{u}_{t,i};\hat{\boldsymbol{\theta}})$.

\begin{figure*}[!ht]
\centering
\includegraphics[width=0.8\textwidth]{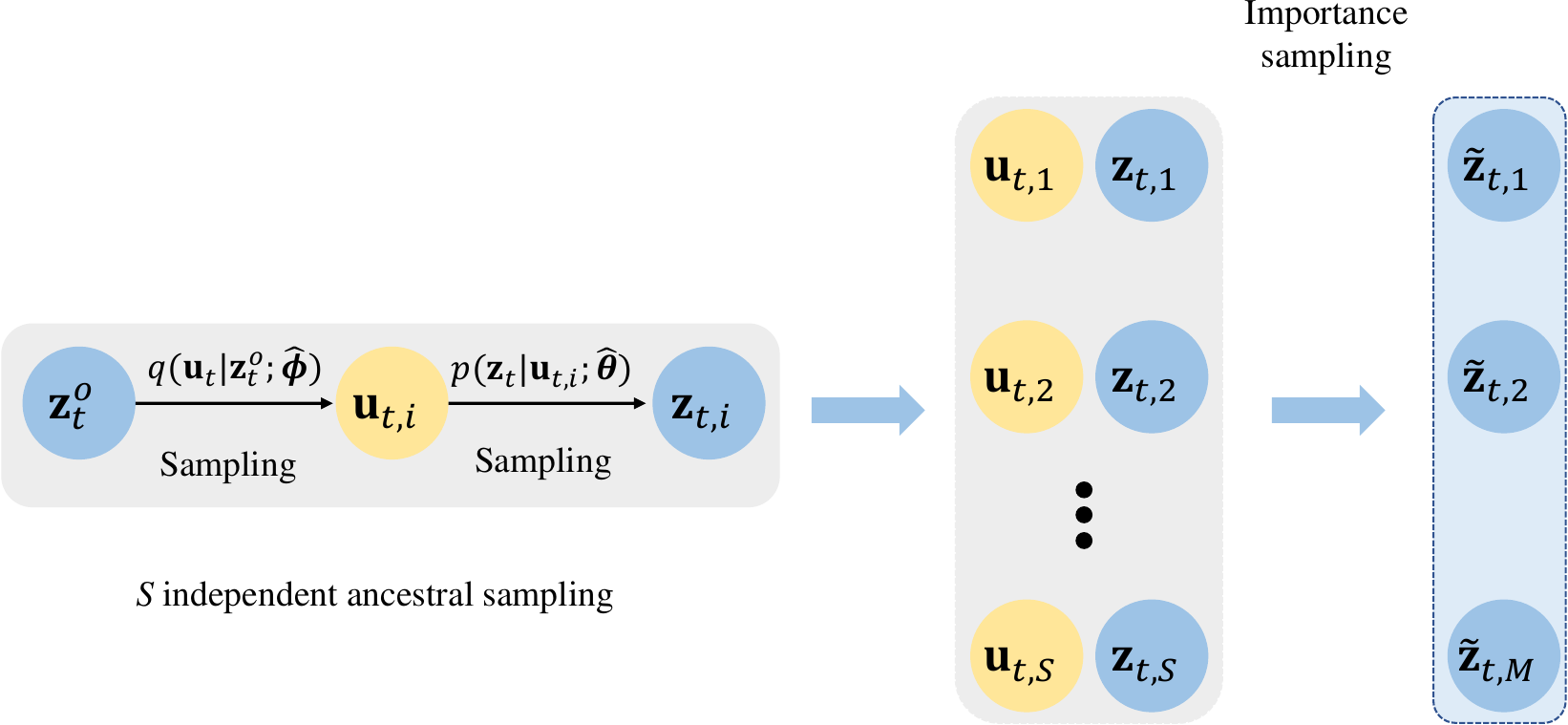}
\caption{Sampling based on the decoder and encoder models for operational forecasting.}
\vspace{-1em}
\label{fig:ir_sampling}
\end{figure*}

Given the pairs $(\mathbf{u}_{t,i},\mathbf{z}_{t,i})$, we reweight them by the
observed-data likelihood. Define
\begin{equation}
\label{eq:weight}
w_i = 
\frac{p\!\left(\mathbf{z}_t^{o}\mid \mathbf{u}_{t,i};{\hat{\boldsymbol{\theta}}}\right)\,p(\mathbf{u}_{t,i})}
     {q\!\left(\mathbf{u}_{t,i}\mid \mathbf{z}_t^o;{\hat{\boldsymbol{\phi}}}\right)},
\qquad
\bar w_i = \frac{w_i}{\sum_{j=1}^{S} w_j}.
\end{equation}
We then resample $M$ indices from $\{1,\dots,S\}$ with probabilities
$\{\bar w_i\}_{i=1}^{S}$ to obtain $\{\tilde{\mathbf{z}}_{t,i}\}_{i=1}^M$.
The predictive scenarios for the target are the last coordinates,
\[
\tilde y_{t+k,i} \;=\; \tilde{\mathbf{z}}_{t,i}[h{+}1], \quad i=1,\dots,M,
\]
which constitutes the communicated probabilistic forecast.
Algorithm~\ref{alg:irs} summarizes the procedure; see Figure~\ref{fig:ir_sampling} for illustration.

\begin{remark}
As noted by \citet{wen2022wind}, Markov chain Monte Carlo based inference requires iterative updates across conditionals for each draw. In contrast, our procedure draws $S$ i.i.d. latents from the variational posterior and performs a single decode per sample, followed by importance resampling. This makes the approach far more efficient operationally.
\end{remark}

\begin{algorithm}
\caption{Importance resampling algorithm}
\label{alg:irs}
\begin{algorithmic}[1]
\Require The estimated encoder $q(\mathbf{u}_t\mid \mathbf{z}_t^o;\hat{\boldsymbol{\phi}})$ and decoder $p(\mathbf{z}_t\mid\mathbf{u}_t;\hat{\boldsymbol{\theta}})$
\Ensure probabilistic scenarios $\{\tilde{\mathbf{z}}_{t,i}\}_{i=1}^M$
\State  Draw $S$ samples $\{\mathbf{u}_{t,i} \}_{i=1}^S$ from the proposal distribution $q(\mathbf{u}_t\mid \mathbf{z}_t^o;\hat{\boldsymbol{\phi}})$
\State For each $\mathbf{u}_{t,i}$, draw a sample $\mathbf{z}_{t,i}$ from $p(\mathbf{z}_t\mid\mathbf{u}_{t,i};\hat{\boldsymbol{\theta}})$ 
\State Compute the importance weights $\Bar{w}_i$ as defined in (\ref{eq:weight}) 
\State Resample $M$ indices from $\{1,\dots,S\}$ with probabilities $\{\bar w_i\}_{i=1}^{S}$ to obtain $\{\tilde{\mathbf{z}}_{t,i}\}_{i=1}^M$
\end{algorithmic}
\end{algorithm}

\section{Case Study}\label{sec:Case}
This section evaluates the proposed method on the U.S. Wind Integration National Dataset (WIND Toolkit).\endnote{https://www.nrel.gov/grid/wind-toolkit.html} The dataset spans 2007–2013 at hourly resolution. We adopt a temporal split: the first 80\% for training and the remaining 20\% for testing. All models are trained on the same training set (with a small held-out validation subset for hyperparameters). We then detail the experimental setup, benchmark models, and validation metrics.
\subsection{Experimental Setups}
Because the original dataset is complete, we inject \emph{synthetic missingness} to evaluate the method. We randomly remove entries to create scattered gaps and vary the missingness rate in $\{5\%,10\%,15\%,25\%\}$ to assess effectiveness (masks are generated with a fixed random seed and independently across time and features). We consider two feature-availability settings:  
\begin{itemize} 
  \item {\verb|Case 1|}: forecasts using only the site’s own data; 
  \item {\verb|Case 2|}: forecasts using the site’s data, along with those from nearby sites. 
\end{itemize}
In both settings, the mask-generation procedure satisfies the MAR assumption by construction. We study lead times $k\in\{1,2,3\}$, choose the input window length via cross-validation, and apply a logit–normal transform to map normalized power values to $\mathbb{R}$ prior to modeling \citep{pinson2012very}; predictions are back-transformed for scoring. 

\subsection{Benchmark Models}
As baselines, we include (i) a naive model that estimates the unconditional forecast distribution from historical samples and (ii) models following the impute-then-predict approach. For context, we also report a reference (“oracle”) model trained on complete data. They are described in detail below:
\begin{itemize}
    \item {\verb|Climatology|}: Forecasts are obtained by estimating the variable’s marginal (unconditional) distribution from past data. Consequently, the forecast does not condition on features and is identical across targets.
    \item {\verb|QR-IM|}: Feature missing values are imputed using the regression-based MissForest method \citep{stekhoven2012missforest} at both training and forecasting. A quantile regression (QR) model is then trained on the imputed training set and applied at test time.
    \item {\verb|Gaussian-IM|}: Feature missing values are imputed with MissForest during both training and operational deployment. A Gaussian forecasting model is then fitted on the imputed training set and applied at test time.
    \item {\verb|DeepAR| \citep{salinas2020deepar}}: We forgo a standalone imputation procedure: missing features are inferred from the recurrent model’s intermediate states and filled in on the fly during model estimation and forecasting.
    \item {\verb|Reference|}: As a reference benchmark, we fit a quantile regression (QR) model using the complete, fully observed dataset.
\end{itemize}
For the proposed model, we vary the number of importance samples $S\in \{1,5,10,20,50 \}$ during training to study the accuracy–cost trade-off. At operational deployment, we use a larger value ($S=10000$) to stabilize importance weights and improve resampling quality.
A desirable property is graceful degradation: under moderate missingness, performance should remain close to the complete-data reference, converging to it as missingness vanishes.

\subsection{Verification Metrics}
The quality of probabilistic forecasts is assessed via sharpness, calibration, and a proper score, i.e., the continuous ranked probability score (CRPS), which are briefly introduced below. For further details, readers are referred to \citet{gneiting2014probabilistic}. 
\begin{itemize}
    \item {\verb|Sharpness|}: The concentration of the predictive distributions. Specifically, it is expressed as the widths of central prediction intervals at several nominal levels.
    \item
    {\verb|Calibration|}: Statistical compatibility of probabilistic forecasts and observations. Here, it is expressed as the empirical coverage of central prediction intervals at several nominal levels.
    \item
    {\verb|CRPS|}: Given the lead time $k$, write $\hat{F}_{t+k}$ the predicted cumulative distribution function, the CRPS is defined as
\begin{equation*}
\begin{aligned}
{\rm CRPS}(\hat{F}_{t+k},y_{t+k})
&= \int_{-\infty}^{\infty} \epsilon_{t+k}(y)^2 \, dy, \\
\epsilon_{t+k}(y)
&:= \hat{F}_{t+k}(y) - \mathbf{1}(y-y_{t+k}).
\end{aligned}
\end{equation*}
where $\mathbf{1}(\cdot)$ is a step function at $y_{t+k}$. We report the average CRPS of all observations in the testing set.
\end{itemize}

\section{Results and Discussion}\label{sec:Results}
In what follows, we report case-wise results and analyze the advantages and caveats of the proposed method.

\subsection{Results}
\subsubsection{Case 1}
\begin{figure}[ht]
\centering
\includegraphics[width=0.9\textwidth]{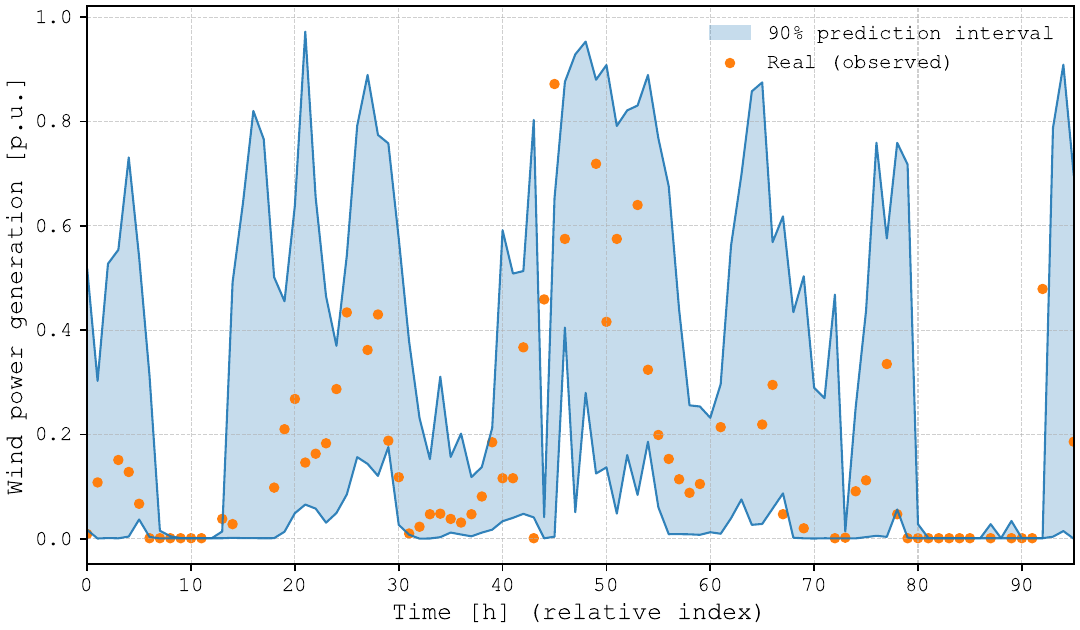}
\caption{4-day episode with 1-step ahead 90\% prediction intervals (issued by our proposed approach), along with corresponding observations.}
\label{fig:pi}
\end{figure}

Table \ref{tab:case1} reports the CRPS values for 1-step-ahead forecasts from the proposed model across missingness rates. As expected, CRPS increases with higher missingness rates. At 5\% missingness rate, performance is close to the complete-data reference (QR; Table \ref{tab:comp}), indicating the effectiveness of the proposed approach. For illustration, Figure~\ref{fig:pi} shows 90\% prediction intervals over a 4-day window at 20\% missingness rate; the intervals cover the observations well, indicating adequate calibration. A head-to-head comparison with the benchmark models follows in the next section.

\begin{table}[!htbp]
\centering
\caption{CRPS for 1-step-ahead forecasts from the proposed model.}
\label{tab:case1}
\setlength{\tabcolsep}{6pt}
{%
\small
\renewcommand{\arraystretch}{0.85}
\begin{tabular*}{0.45\linewidth}{@{\extracolsep{\fill}} l *{5}{S[table-format=2.1]} @{}}
\toprule
\multirow{2}{*}{\textbf{Metric}} & \multicolumn{5}{c}{\textbf{Missingness rate (\%)}} \\
\cmidrule(lr){2-6}
 & {5} & {10} & {15} & {20} & {25} \\
\midrule
\textbf{CRPS} & 6.9 & 7.0 & 7.2 & 7.3 & 7.5 \\
\bottomrule
\end{tabular*}
}

\vspace{0.5ex}
{\raggedright\footnotesize
\textit{Notes:} Lower is better. Values are percentages of plant capacity (normalized); 
Lead time $k{=}1$.\par}
\end{table}

\subsubsection{Case 2}
We now assess forecasts at the target site when its own data have a 20\% missingness rate, while two neighboring sites are used as auxiliary features (AFs) with varying levels of missingness. Table \ref{tab:case2} reports the CRPS values for 1-step-ahead forecasts; the percentages denote the missingness rate of the AFs. Relative to using no AFs (CRPS =7.3; Table \ref{tab:case1}), incorporating neighbors consistently improves forecast quality, with the largest gain when AFs are fully available and persistent benefits even when AFs exhibit 20\% missingness. These results indicate that the proposed model effectively exploits additional relevant information and remains useful under partial availability by marginalizing missing features rather than relying on pre-imputation.

\begin{table}[!htbp]
\centering
\caption{CRPS for 1-step forecasts with different missingness rates on auxiliary features (AFs).}
\label{tab:case2}

{%
\small
\setlength{\tabcolsep}{6pt}%
\renewcommand{\arraystretch}{0.85}%
\begin{tabular*}{0.4\linewidth}{@{\extracolsep{\fill}} c *{4}{S[table-format=2.1]} @{}}
\toprule
\multirow{2}{*}{\textbf{Metric}} & \multicolumn{4}{c}{\textbf{AFs missingness rate (\%)}} \\
\cmidrule(lr){2-5}
 & {0} & {5} & {10} & {20} \\
\midrule
\textbf{CRPS} & 6.9 & 7.0 & 7.1 & 7.1 \\
\bottomrule
\end{tabular*}
}

\vspace{0.5ex}
{\raggedright\footnotesize
\textit{Notes:} Lower is better. Values are percentages of plant capacity (normalized); 
lead time $k{=}1$.\par}
\end{table}

\subsection{Comparison with Benchmarks}

\begin{table*}[!t]
\centering
\caption{CRPS at 20\% missingness rate for different lead times.}
\label{tab:comp}

{%
\small
\setlength{\tabcolsep}{5pt}%
\renewcommand{\arraystretch}{1.05}%
\begin{tabular*}{\textwidth}{@{\extracolsep{\fill}} c *{6}{S[table-format=2.1]} @{}}
\toprule
\textbf{Lead time} &
\textbf{Climatology} & \textbf{QR-IM} & \textbf{Gaussian-IM} &
\textbf{DeepAR} & \textbf{Reference} & \textbf{Proposed} \\
\midrule
1 & 18.6 & 7.8 & 7.6 & 7.8 & 6.9 & 7.3 \\
2 & 18.6 & 10.1 & 10.2 & 10.2 & 9.3 & 9.7 \\
3 & 18.6 & 11.9 & 11.9 & 12.1 & 11.2 & 11.5 \\
\bottomrule
\end{tabular*}
}

\vspace{0.5ex}
{\raggedright\footnotesize
\textit{Notes:} Lower is better. Values are percentages of plant capacity (normalized).\par}
\end{table*}

For comparison, Table~\ref{tab:comp} reports CRPS values for Case 1 at a 20\% missingness rate. Overall, the DeepAR implementation underperforms compared to both other impute-then-predict baselines and our joint-modeling approach. Unlike our method, which estimates parameters by maximizing the observed-data likelihood and marginalizing missing features, DeepAR imputes missing values using RNN hidden states and then applies standard estimation. In this sense, it is an effective impute-then-predict procedure performed on the fly. However, this procedure neither optimizes the observed-data likelihood nor explicitly propagates missing-feature uncertainty, which helps explain why its quality rivals that of even simple models trained on imputed data. While such end-to-end sequence models reduce manual preprocessing, their ad-hoc handling of missing data can be statistically unprincipled and, in our experiments, results in inferior forecast skill.

The impute-then-predict baselines generally underperform our joint-modeling approach. Although the principle is intuitive (i.e., accurate imputations should yield small prediction errors), imputation inevitably introduces model-misspecification and estimation bias within the forecasting pipeline, even with strong imputers such as MissForest (used in QR-IM and Gaussian-IM). In contrast, the proposed model maximizes the observed-data likelihood and marginalizes missing features; under MAR assumptions, this yields consistent parameter estimates without explicitly modeling the missingness mechanism.

\begin{figure}[!ht]
\centering

\begin{subfigure}[b]{0.9\linewidth}
    \centering
    \includegraphics[width=0.75\linewidth]{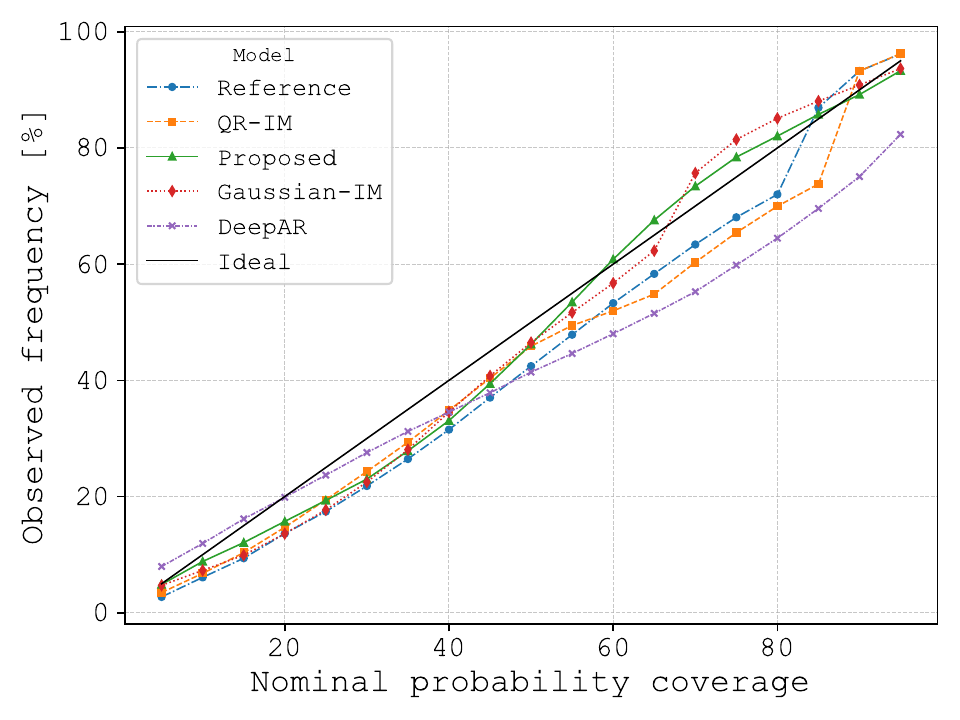}
    \caption{Reliability diagrams for 1-step forecasts.}
    \label{fig:reliability}
    \vspace{-1em}
\end{subfigure}

\medskip 

\begin{subfigure}[b]{0.9\linewidth}
    \centering
    \includegraphics[width=0.75\linewidth]{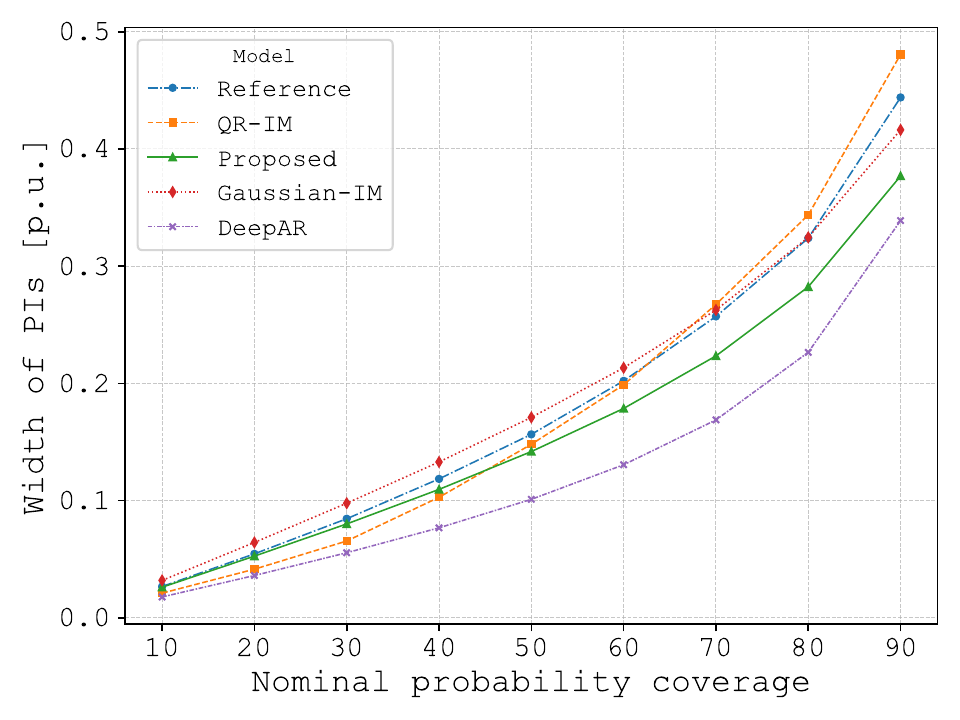}
    \caption{Sharpness diagrams for 1-step forecasts.}
    \label{fig:sharpness}
\end{subfigure}

\caption{Reliability and sharpness diagrams for 1-step forecasts.}
\label{fig:rel_sharp}
\vspace{-1em}
\end{figure}

Figures~\ref{fig:reliability} and \ref{fig:sharpness} compare reliability and sharpness for 1-step forecasts. Reliability diagrams for Gaussian-IM and our method are both close to the reference model, indicating good calibration; however, our method delivers sharper predictive distributions than Gaussian-IM, reflecting better uncertainty concentration at similar calibration. We note that DeepAR produces the sharpest predictive distributions. This is likely a consequence of its on-the-fly impute-then-predict mechanism, which tends to underestimate variance. In particular, we have
\begin{equation*}
\begin{aligned}
\mathrm{Var}(Y_{t+k}\mid \mathbf{y}_t^o)
&= \mathbb{E}\!\big[\mathrm{Var}(Y_{t+k}\mid \mathbf{y}_t^o,\mathbf{y}_t^m)\big] \\
&\quad
 + \mathrm{Var}\!\big(\mathbb{E}[Y_{t+k}\mid \mathbf{y}_t^o,\mathbf{y}_t^m]\big).
\end{aligned}
\end{equation*}

Conditioning on a single point imputation treats the imputed values as error-free and effectively removes the variance across plausible imputations (the second term in the conditional law of total variance). This leads to under-dispersed forecasts, i.e., narrow prediction intervals and optimistic uncertainty relative to approaches that marginalize over the missing features.

\subsection{Component and Hyperparameter Analysis}
\subsubsection{Effect of the Number of Importance Samples \texorpdfstring{($S$)}{(S)} on the ELBO}
As in \eqref{eq:elbo}, the importance-weighted lower bound is estimated using $S$ samples from the latent posterior distribution. We assess the effect of $S$ in Case 1 at a 20\% missingness rate with lead time $k=1$, varying $S\in \{1, 5, 10, 20, 50\}$. Table~\ref{tab:K} reports the resulting CRPS values for 1-step forecasts. Empirically, larger $S$ generally reduces CRPS, with diminishing returns beyond $S \simeq 20$. This aligns with \citet{burda2015importance}, where increasing the number of importance samples tightens the IWAE bound, yielding a closer approximation to the log-likelihood and more accurate parameter estimates. The trade-off is computational: both training and inference costs grow in $S$.

\begin{table}[!htbp]
\centering
\caption{CRPS for 1-step-ahead forecasts from the proposed model with varying $S$ during training.}
\label{tab:K}

{%
\small
\setlength{\tabcolsep}{6pt}%
\renewcommand{\arraystretch}{1.0}
\begin{tabular*}{0.45\linewidth}{@{\extracolsep{\fill}} c *{5}{S[table-format=2.1]} @{}}
\toprule
\multirow{2}{*}{\textbf{Metric}} &
\multicolumn{5}{c}{\textbf{$S$ (importance samples)}} \\
\cmidrule(lr){2-6}
 & {1} & {5} & {10} & {20} & {50} \\
\midrule
\textbf{CRPS} & 7.8 & 7.5 & 7.4 & 7.4 & 7.3 \\
\bottomrule
\end{tabular*}
}

\vspace{0.5ex}
{\raggedright\footnotesize
\textit{Notes:} Lower is better. Values are percentages of plant capacity (normalized);
lead time $k{=}1$.\par}
\end{table}

\subsubsection{Posterior Approximation: Impact on Forecasts}

Rather than restricting the variational posterior to a fixed family, we approximate it with a normalizing flow model, which provides greater flexibility. For the ELBO (i.e., the $S=1$ case of \eqref{eq:elbo}), the bound can be written as
\begin{equation*}
\begin{aligned}
\sum_{t=1}^T &\Big(
   -\mathrm{KL}\big[q(\mathbf{u}_t\mid \mathbf{z}_t^o;\boldsymbol{\phi})
      \,\|\, p(\mathbf{u}_t)\big] \\
&\qquad
   +\, \mathbb{E}_{q(\mathbf{u}_t \mid \mathbf{z}_t^o;\boldsymbol{\phi})}
      \big[\log p(\mathbf{z}_t^o\mid\mathbf{u}_t;\boldsymbol{\theta})\big]
\Big),
\end{aligned}
\end{equation*}
where $\mathrm{KL}(\cdot)$ represents the Kullback–Leibler divergence. The bound is tight when the variational posterior $q(\mathbf{u}_t\mid \mathbf{z}_t^o;\boldsymbol{\phi})$ matches the true posterior. For comparison, in Case 1 at a 20\% missingness rate, we also use a Gaussian variational posterior in place of the flow; this yields a CRPS of 8.1 for 1-step forecasts, which is higher (worse) than that of our proposed model.

\subsection{Training time}

The training cost of the proposed model is independent of the number of scenarios required for operational deployment and is less sensitive to feature dimensions. Table~\ref{tab:time} reports training times for Case 1, where the neural network based models are trained over 100 epochs; the proposed model trains within practical limits. In Case 2, where the feature dimension is roughly 3 times that of Case 1, the training time increases only marginally, reflecting the model’s mild scaling with input size. At operation, scenarios are generated via ancestral sampling, followed by importance resampling—a non-iterative procedure that avoids the repeated updates typical of Markov Chain Monte Carlo or chained imputations.

\begin{table}[!htbp]
\centering
\caption{Training time for each model in Case 1 (minutes).}
\label{tab:time}
\small
\setlength{\tabcolsep}{6pt}
\begin{tabular*}{0.7\linewidth}{@{\extracolsep{\fill}} c *{4}{S} @{}}
\toprule
\textbf{Metric} & \multicolumn{1}{c}{\textbf{QR-IM}}
                & \multicolumn{1}{c}{\textbf{Gaussian-IM}}
                & \multicolumn{1}{c}{\textbf{DeepAR}}
                & \multicolumn{1}{c}{\textbf{Proposed}} \\
\midrule
\textbf{Time (min)} & 1 & 32 & 67 & 10 \\
\bottomrule
\end{tabular*}
\end{table}

\section{Concluding Remarks}\label{sec:Conclusion}
Compared with the impute-then-predict approach, the joint modeling method estimates parameters by maximizing the observed-data likelihood under the MAR mechanism (ignorability). When the model is well specified, especially with expressive generative architectures, this yields more skillful and better-calibrated forecasts than pipelines that first impute and then predict.
In this paper, we (i) explain why impute-then-predict biases parameter estimates and fails to propagate missing-feature uncertainty (conditioning on point imputations drops the variance across plausible imputations), and (ii) introduce an efficient generative method that improves forecast quality. The method is operationally simple: it generates several probabilistic scenarios via ancestral sampling with importance resampling, without iterative chains, making it well suited for real-time decision-making under uncertainty.

Our study models the data distribution within fixed-length windows, which simplifies estimation but ignores temporal dependence and the sequential structure of time series. A natural extension is to incorporate sequence models (e.g., state-space or latent-dynamics VAEs/flows) that update the predictive distribution recursively across time and capture longer-range dependencies.
In practice, missingness may be MNAR, requiring models that jointly specify the data and the missingness mechanism. The proposed generative framework lends itself to such extensions by modeling $p(\mathbf{z},\mathbf{m})=p(\mathbf{z})p(\mathbf{m}\mid \mathbf{z})$ and optimizing the corresponding full likelihood. Future work should explore identifiability conditions, practical parameterizations of $p(\mathbf{m}\mid \mathbf{z})$, and sensitivity analysis when MNAR assumptions are only partially credible.
More generally, methods that maintain predictive validity under incomplete observations are fundamental to the development of robust, data-driven decision-making procedures for power system operations and for applications characterized by uncertainty and missing data.

\begingroup \parindent 0pt \parskip 0.0ex \def\enotesize{\normalsize} \theendnotes \endgroup

%
%
%

\ACKNOWLEDGMENT{This work is funded by the National Natural Science Foundation of China (52307119). The author appreciates the constructive suggestions by anonymous reviewers.}



\bibliographystyle{informs2014}  
\bibliography{reference}          




\end{document}